\documentclass[11pt]{article}

\usepackage[]{acl}

\usepackage{times}
\usepackage{latexsym}

\usepackage[T1]{fontenc}

\usepackage[utf8]{inputenc}

\usepackage{microtype}

\usepackage{mathtools}
\usepackage{graphics}
\usepackage{graphicx}
\usepackage{amsmath}
\usepackage{amsfonts,amssymb}
\usepackage{amssymb}
\usepackage{bbm}
\usepackage[ruled, linesnumbered]{algorithm2e}
\usepackage{amsmath}
\usepackage{booktabs}
\usepackage{multirow}
\usepackage{color}
\usepackage{xcolor}
\usepackage{pifont}
\usepackage{pgfplots}
\usepackage{makecell}
\usepackage{caption}
\usepackage{subcaption}
\usepackage{math_cmd}

\newcommand{\heart}{\text{\small \ding{170}}}

\title{GNN-SL: Sequence Labeling Based on Nearest Examples via GNN}

\author{
Shuhe Wang$^{\spadesuit}$*,
Yuxian Meng$^{\clubsuit}$,
Rongbin Ouyang$^{\spadesuit}$,
Jiwei Li$^{\blacklozenge}$\\
{\bf Tianwei Zhang$^{\heart}$,
Lingjuan Lyu$^{\blacktriangle}$,
Guoyin Wang$^{\bigstar}$}\dag}

\begin{document}
\maketitle

\begingroup\def\thefootnote{*}\footnotetext{wangshuhe@stu.pku.edu.cn}\endgroup
\begingroup\def\thefootnote{\dag}\footnotetext{$^{\spadesuit}$ Peking University, $^{\clubsuit}$ JQ Investments, $^{\blacklozenge}$ Zhejiang University, $^{\heart}$ Nanyang Technological University, $^{\blacktriangle}$ Sony AI, $^{\bigstar}$ Amazon}\endgroup

\begin{abstract}
To better handle long-tail cases in the sequence labeling (SL) task, in this work, we introduce graph neural networks sequence labeling (GNN-SL), which augments the vanilla SL model output with similar tagging examples retrieved from the whole training set. Since not all the retrieved tagging examples benefit the model prediction, we construct a heterogeneous graph, and leverage graph neural networks (GNNs) to transfer information between the retrieved tagging examples and the input word sequence. The augmented node which aggregates information from neighbors is used to do prediction. This strategy enables the model to directly acquire similar tagging examples and improves the 
 general 
 quality of predictions. We conduct a variety of experiments on three 
 typical 
 sequence labeling tasks: Named Entity Recognition (NER), Part of Speech Tagging (POS), and Chinese Word Segmentation (CWS) to show the significant performance of our GNN-SL. Notably, GNN-SL achieves SOTA results of 96.9 (+0.2) on PKU, 98.3 (+0.4) on CITYU, 98.5 (+0.2) on MSR, and 96.9 (+0.2) on AS for the CWS task, and results
comparable to SOTA performances 
on NER datasets, and POS datasets.\footnote{Code is available at \url{https://github.com/ShuheWang1998/GNN-SL}. }
\end{abstract}

\section{Introduction}
Sequence labeling (SL) is a fundamental problem in NLP, which encompasses a variety of tasks e.g., Named Entity Recognition (NER), Part of Speech Tagging (POS), and Chinese Word Segmentation (CWS). Most existing sequence labeling algorithms  \cite{clark2018semi,zhang2018chinese,bohnet2018morphosyntactic,shao2017character,meng2019glyce} can be decomposed into two parts: (1) 
representation learning: 
mapping each input word to a higher-dimensional contextual vector using neural network models such as LSTMs \cite{huang2019toward}, CNNs \cite{wang2020astral}, or pretrained language models \cite{devlin2018bert};
and (2) 
classification: fitting the vector representation of each word to a softmax layer to obtain the classification label.

\begin{figure}[htb]
\centering
    \includegraphics[scale=0.26]{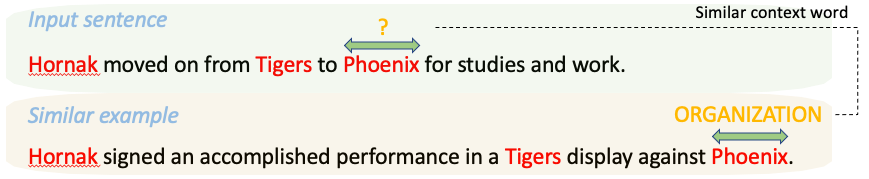}
    \caption{Example for the NER assignment when given a similar example.}
    \label{fig:intro_example}
\end{figure}
Because
the protocol described above 
relies on the model's ability to memorize the characteristics of training examples,
its performance plummets when handling long-tail cases or minority categories: 
it is hard for models to 
memorize long-tail cases with a relatively small number of gradient updates during training. 
 Intuitively, it's easier for a model to make predictions on long-term cases
 at test time when it is able to refer to similar training examples. 
  For example,  in Figure \ref{fig:intro_example}, the model can more easily label the word “{\it Phoenix}” in the given sentence “{\it Hornak moved on from Tigers to Phoenix for studies and work}” as an “{\it ORGANIZATION}” entity when referring to a similar example “{\it Hornak signed an accomplished performance in a Tigers display against Phoenix}”. 
  
Benefiting from the success of augmented models in NLP  \cite{khandelwal2019generalization,khandelwal2020nearest,guu2020realm,lewis2020pre,meng2021gnn} 
, a simple yet effective method to mitigate the above issues is to apply  
the $k$ nearest neighbors ($k$NN) strategy: The $k$NN model retrieves $k$ similar tagging examples from a large cached datastore for each input word and augments the prediction with 
the probability computed by the cosine similarity between the input word and each of the retrieved nearest neighbors. 
Unfortunately, there is a significant shortcoming of this strategy.
Retrieved neighbors are related to the input word in different ways:
some are related in semantics while others in syntactic, 
some are close to the original input word while others are just noise. 
A more sophisticated model is required to model the relationships between retrieved examples and the input word.

In this work, inspired by recent progress in combining graph neural networks (GNNs) with augmented models \cite{meng2021gnn}, we propose GNN-SL to provide a general sequence-labeling model with the ability of 
 effectively 
referring to training examples at test time. 
The core idea of GNN-SL is to build a graph between the retrieved nearest training examples and the input word, and use graph neural networks (GNNs) to model their
relationships. 
To this end,
we construct an undirected graph, where nodes represent both the input words and retrieved training examples, 
 and edges represent the relationship between each node.
The message is passed between the input words and retrieved training examples. In this way, we are able to more effectively harness evidence from the retrieved neighbors 
in the training set
and by aggregating information from them, 
better token-level representations are obtained 
for final predictions.


To evaluate the effectiveness of GNN-SL, we conduct experiments over three 
widely-used
sequence labeling tasks: Named Entity Recognition (NER), Part of Speech Tagging (POS), and Chinese Word Segmentation (CWS), and choose both English and Chinese datasets as benchmarks. Notably, applying the GNN-SL to the ChineseBERT \cite{sun2021chinesebert}, a Chinese robust pre-training language model, we achieve SOTA results of 96.9 (+0.2) on PKU, 98.3 (+0.4) on CITYU, 98.5 (+0.2) on MSR, and 96.9 (+0.2) on AS for the CWS task.
We also achieve performances comparable to current SOTA results on CoNLL, OntoNotes5.0, OntoNotes4.0 and MSRA for NER, and CTB5, CTB6, UD1.4, WSJ and Tweets for  POS. 
We also conduct comprehensive ablation experiments to better understand the working mechanism of GNN-SL.

\section{Related Work}
\paragraph{Sequence Labeling} Sequence labeling (SL) encompasses a variety of NLP tasks e.g., Named Entity Recognition (NER), Part of Speech Tagging (POS), and Chinese Word Segmentation (CWS). With the development of machine learning,  neural network models have been widely used as the backbone for the sequence labeling task. For example, \citet{hammerton2003named} use unidirectional LSTMs to solve the NER task, \citet{collobert2011natural} and \citet{lample2016neural} combine CRFs with CNN and LSTMs respectively. To extract fine-grained information of words, \citet{ma2016end} and \citet{chiu2016named} add character features via character CNN. \citet{liu2018empower,lin2021asrnn,cui2019hierarchically} focus on the decoder including more context information into the decoding word. Recently there have been several efforts to optimize the SL task from different views: interpolating latent variables \cite{lin2020enhanced,shao2021self}; combining positive information \cite{dai2019joint}; viewing it as a machine reading comprehension (MRC) task \cite{li2019unified,li2019dice,gan2021dependency}.

\paragraph{Retrieval Augmented Model} Retrieval augmented models additionally use the input to retrieve information from the constructed datastore to the model performance. As described in \citet{meng2021gnn}, this process can be understood as {\it “an open-book exam is easier than a close-book exam”}. The retrieval augmented model is more familiar in the question answering task, in which the model generates related answers from a constructed datastore \cite{karpukhin2020dense,xiong2020approximate,yih2020retrieval}. Recently other NLP tasks have introduced this approach and achieved a good performance, such as language modeling (LM) \cite{khandelwal2019generalization, meng2021gnn}, dialog generation \cite{fan2020augmenting,thulke2021efficient}, neural machine translation (NMT) \cite{khandelwal2020nearest,meng2021fast,wang2021faster}.

\paragraph{Graph Neural Networks} The key idea behind graph neural networks (GNNs) is to aggregate feature information from the local neighbors of the node via neural networks \cite{liu2018heterogeneous,velivckovic2017graph,hamilton2017inductive}. Recently more and more researchers have proved the effectiveness of GNNs in the NLP task. For text classification, \citet{yao2019graph} uses a Text Graph Convolution Network (Text GCN) to learn the embeddings for both words and documents on a graph based on word co-occurrence and document word relations. For question answering, \citet{song2018exploring} performs evidence integration forming more complex graphs compared to DAGs, and \citet{de2018question} frames the problem as an inference problem on a graph, in which mentions of entities are nodes of this graph while edges encode relations between different mentions. For information extraction, \citet{lin2020enhanced} characterizes the complex interaction between sentences and potential relation instances via a graph-enhanced dual attention network (GEDA). For the recent work GNN-LM, \citet{meng2021gnn} builds an undirected heterogeneous graph between an input context and its semantically related neighbors selected from the training corpus, GNNs are constructed upon the graph to aggregate information from similar contexts to decode the token.

\begin{figure*}[htb]
    \includegraphics[scale=0.384]{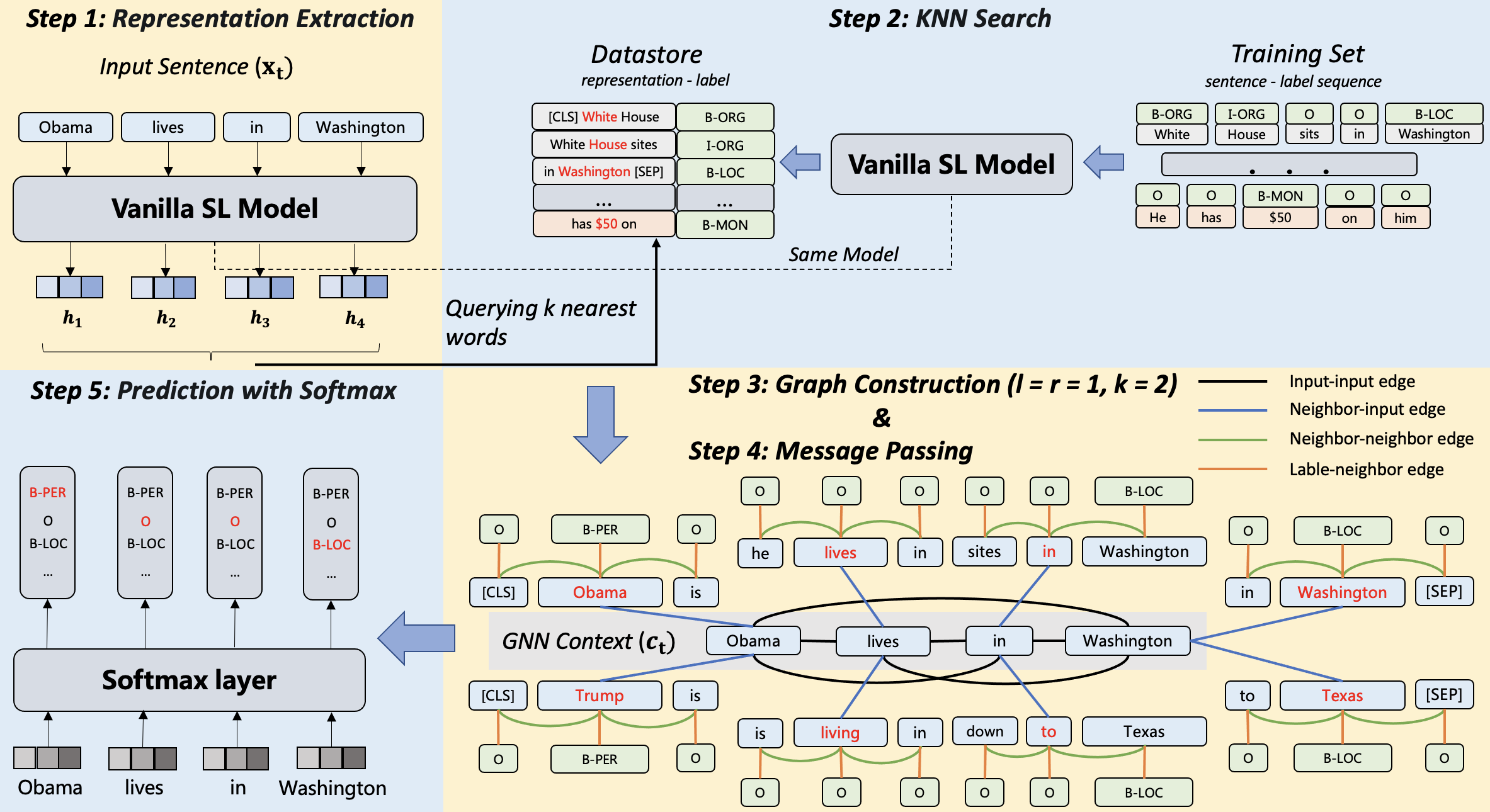}
    \caption{An example for the process of GNN-SL. \text{\bf{Step 1 Representation Extraction}}: Suppose that we need to extract named entities for the given sentence: \textit{Obama lives in Washington}. The representation for each word is the 
     last hidden state of the pretrained vanilla sequence labeling model. 
     \text{\bf{Step 2 $k$NN Search}}: Obtaining $k$ nearest neighbors for \textsc{each} input word in the cached datastore, which consists of representation-label pairs for all training data.
        \text{\bf{Step 3\&4 Graph Construction \& Message Passing}}: The queried $k$ nearest neighbors and the input word are constructed into a graph. 
        The message is passed from the nearest neighbors to each input word to obtain the aggregated representation. 
        \text{\bf{Step 5 Prediction with Softmax}}: The aggregated representation of each word is passed to a softmax layer to compute the likelihood of the assigned label.}
    \label{fig:process}
\end{figure*}

\section{$k$NN-SL}
\label{knn_method}
Sequence labeling (SL) is a typical NLP task, which assigns a label $y\in Y$ to each word $w$ in the given input word sequence $x = \{w_{1},\dots,w_{n}\}$, where $n$ denotes the length of the given sentence. 
We assume that $\{\mathcal{X}, \mathcal{Y}\}=\{(x^{1}, y^{1}),\dots,(x^{N}, y^{N})\}$ denotes the training set, where $(x^{i}, y^{i}), \forall 1\leq i \leq N$ 
denotes the pair containing 
a word sequence and its corresponding label sequence. Let $N$ be the size of the training set. 

\subsection{$k$NN-SL} 
\label{knn_sl}
The key idea of the $k$NN-SL model is 
to augment
 the process of classification during the inference stage with a $k$ nearest neighbor retrieval mechanism, which can be split into the following pipelines: (1) 
 using an already-trained sequence labeling model (e.g., BERT \cite{devlin2018bert} or RoBERTa \cite{liu2019roberta})
 to obtain 
  word representation $h$ for each token with the input word sequence; 
  (2) using $h$ as the query and finding the most similar $k$ tokens in the cached datastore which is constructed by the training set; and
  (3) augmenting the classification probability generated by the vanilla SL model (i.e., $p_{\text{vanilla}}$) with the 
  $k$NN label distribution $p_{\text{kNN}}$ to obtain the final distribution.

\paragraph{Vanilla probability $p_{\text{vanilla}}$} For a given word $w$, the output $h$ generated from the last layer of the vanilla SL model is used as its representation, where $h\in\mathbb{R}^{m}$. Then  $h$ is fed into a multi-layer perceptron (MLP) to obtain the probability distribution $p_{\text{vanilla}}$ via a softmax layer: 
\begin{equation}
    \begin{aligned}
    	p_{\text{vanilla}}(y|w,x) = \text{softmax}(\text{MLP}(h))
    \end{aligned}
\end{equation}

\paragraph{$k$NN-augmented probability $p_{\text{kNN}}$.} For each word $w$, its corresponding embedding $h$ is used to query $k$ nearest neighbors set $\mathcal{N}$ from the training set using $L^{2}$ Euclidean distance $d(h, \cdot)$ as similarity measure. The retrieved nearest neighbors set $\mathcal{N}$ is formulated as $(key, value)$ pairs, where $key$ represents the retrieved similar word and $value$ represents the corresponding SL label.

Then the retrieved examples are converted into a distribution over the label vocabulary based on an RBF kernel output \cite{vert2004primer} of the distance to the original embedding $h$:

\begin{equation}
    \begin{aligned}
    	p_{\text{kNN}}(y|w,x) \propto \sum_{(k,v)\in\mathcal{N}}\mathbbm{1}_{y=v}\exp(\frac{-d(k,h)}{T})
    \end{aligned}
\end{equation}
where $T$ is a temperature parameter to flatten the distribution. Finally, the vanilla distribution $p_{\text{vanilla}}(y|w,x)$ is augmented with $p_{\text{kNN}}(y|w,x)$ generating the final distribution $p_{\text{final}}(y|w,x)$:

\begin{equation}
\label{knn_formulation}
    \begin{aligned}
    	p_{\text{final}}(y|w,x) = &\lambda p_{\text{vanilla}}(y|w,x) +\\ &(1-\lambda)p_{\text{kNN}}(y|w,x)
    \end{aligned}
\end{equation}
where $\lambda$ is adjustable to make a balance between $k$NN distribution and vanilla distribution.

\section{GNN-SL}
\label{proposed_method}

\subsection{Overview}
Intuitively, 
retrieved neighbors are related to the input word in different ways: 
some are similar in semantics while others in syntactic;
some are very similar to the input word while others are just noise. 
To better model the relationships between retrieved neighbors and the input word, 
we propose graph neural networks sequence labeling (GNN-SL).

The proposed GNN-SL can be decomposed into five steps: (1) 
 obtaining token features 
  using a pre-trained vanilla sequence labeling model, which is the same as in KNN-SL; (2) obtaining $k$ nearest neighbors from the whole training set for each input word; 
    (3) constructing an undirected graph between each word within the sentence and its $k$ nearest neighbors; (4) 
obtaining aggregated word representations through messages passing along the graph; and (5)
feeding the aggregated word representation to the softmax layer to obtain the final label. 
The full pipeline is shown in Figure \ref{fig:process}.

For steps (1) and (2), they are akin the strategies taken in KNN-SL.
We will describe the details for steps (3) and (4) in order below. 



\subsection{Graph Construction}
\label{building_graph}
We formulate the graph as $\mathcal{G}=(\mathcal{V},\mathcal{E},\mathcal{A},\mathcal{R})$, where $\mathcal{V}$ represents a collection of nodes $\mathit{v}$ and $\mathcal{E}$ represents a collection of edges $\mathit{e}$. 
$\mathcal{A}$ refers to node types
 and $\mathcal{R}$ refers to edge types.
\paragraph{Nodes}
In the constructed graph, we define three types of nodes $\mathcal{A}$:

(1)  {\bf Input} nodes, denoted by $a_{\text{input}}\in \mathcal{A}$, which correspond to words of the input sentence. 
In the example of Figure \ref{fig:process} Step 3, the input nodes are displayed with the word sequence $x = \{\text{Obama}, \text{lives}, \text{in}, \text{Washington}\}$;


(2)  {\bf Neighbor} nodes, denoted by $a_{\text{neighbor}}\in \mathcal{A}$, which correspond to words in the retrieved neighbors. 
The context of nearest neighbors is also included (and thus treated as neighbor nodes)  in an attempt to capture more abundant contextual information for the retrieved neighbors.
For the example in Figure  \ref{fig:process}, 
for each input word with representation $h$, $k$ nearest neighbors are queried from the cached representations of all words in the training set with the $L2$ distance as the metric of similarity. Taking the input word $\{\text{Obama}\}$ as the example with $k=2$,
we obtain two nearest neighbors $\{{\text{Obama}, \text{Trump}}\}$ 
leveraging the $k$NN search. 
The contexts of each retrieved nearest neighbor are also considered by adding both left and right contexts around the retrieved nearest neighbor, where
$\{{\text{Obama}}\}$ is expanded to $\{\text{[CLS]}, \text{Obama}, \text{is}\}$ 
and 
$\{{\text{Trump}}\}$ is expanded to $\{\text{[CLS]}, \text{Trump}, \text{is}\}\}$\footnote{\text{[CLS]} is a special token usually applied in pre-training language models (e.g., BERT) representing the beginning of a sentence.}. 
 The analysis of the size of the context is conducted in Section \ref{ablation_study}.

(3)  {\bf Label} nodes,  denoted by $a_{\text{label}}\in \mathcal{A}$, 
since the labels of nearest neighbors provide important evidence for the input node to classify, we wish to pass the influence of neighbors' labels to the input node along the graph. 
As will be shown in ablation studies in Section \ref{ablation_study}, the consideration of label nodes introduces a significant performance boost. 
Shown in  Figure \ref{fig:process} Step 3, taking the input word $\{\text{Obama}\}$ as the example, the two retrieved nearest neighbors are $\{{\text{Obama}}\}$ and $\{{\text{Trump}}\}$, and both the corresponding node labels are $\{\text{B-PER}\}$.

With the above formulated, 
$\mathcal{A}$ can be rewritten as $\{a_{\text{input}}, a_{\text{neighbor}}, a_{\text{label}}\}$.

\paragraph{Edges} Given the three types of nodes $\mathcal{A}=\{a_{\text{input}}, a_{\text{neighbor}}, a_{\text{label}}\}$, we connect them 
using different types of edges to enable
 information passing.

We define four types of edges for $\mathcal{R}$: (1) edges within the input nodes $a_{\text{input}}$, notated by $r_{\text{input-input}}$; (2) edges between the neighbor nodes $a_{\text{neighbor}}$ and the input nodes $a_{\text{input}}$, notated by $r_{\text{neighbor-input}}$; (3)  edges within the neighbor nodes $a_{\text{neighbor}}$, notated by $r_{\text{neighbor-neighbor}}$; and (4) edges between the label nodes $a_{\text{label}}$ and the neighbor nodes $a_{\text{neighbor}}$, denoted by $r_{\text{label-neighbor}}$.
All types of edges are bi-directional which allows information  passing on both sides.
 We use different colors to differentiate different relations in Figure \ref{fig:process} Step 3.

For $r_{\text{input-input}}$ and $r_{\text{neighbor-neighbor}}$, they respectively mimic the attention mechanism to aggregate the context information within the input word sequence or the expanded nearest context, which are shown with the black and green color in Figure \ref{fig:process}. For $r_{\text{neighbor-input}}$, it connects the retrieved neighbors and the query input word,  transferring the neighbor information to the input word. For $r_{\text{label-neighbor}}$ colored with orange, information is passed from label nodes to neighbor nodes,  which is ultimately transferred to input nodes.


\subsection{Message Passing On The Graph}
Given the constructed graph, we next use  
 graph neural networks (GNNs) to aggregate information based on the graph to obtain the final representation for each token to classify. 
More formally, we define the l-th layer representation of node $n$ as follows:
\begin{equation}
    \begin{aligned}
        h_{n}^{l} = \mathop{\text{Aggregate}}\limits_{\forall s\in \mathcal{N}(n)}(&\text{A}(s,e,n)\cdot\text{M}(s,e,n)) + h_{n}^{l-1}
    \end{aligned}
\end{equation}
where $\text{M}(s,e,n)$ denotes the information transferred from the node $s$ to the node $n$ along the edge $e$, $\text{A}(s,e,n)$ denotes the edge weight modeling the importance of the source node $s$ on the target node $n$ with the relationship $e$, and $\text{Aggregate}(\cdot)$ denotes the function to aggregate the transferred information from the neighbors of node $n$. We detail how to obtain A($\cdot$), M($\cdot$), and Aggregate($\cdot$) below.


\paragraph{Message} For each edge $(s,e,n)$, the message transferred from the source node $s$ to the target node $n$ can be formulated as:
\begin{equation}
    \begin{aligned}
    	\text{M}(s,e,n) = W^{v}_{\tau(s)}h^{l-1}_{s}W_{\phi(e)}
    \end{aligned}
\end{equation}
where $d$ denotes the dimensionality of the vector, $W^{v}_{\tau(s)}\in \mathbb{R}^{d\times d}$ and $W_{\phi(e)}\in \mathbb{R}^{d\times d}$ are two learnable weight matrixes controling the outflow of node $s$ from the node side and the edge side respectively.

For the receiver node $n$, the importance of each neighbor $s$ with the relationship $e$ can be viewed as the attention mechanism in \newcite{vaswani2017attention}, which relates different positions of a single sequence. To introduce the attention weights, we first map the source node $s$ to a key vector $K(s)$ and the target node $n$ to a query vector $Q(n)$:
\begin{equation}
    \begin{aligned}
    	K(s)=W^{k}_{\tau(s)}h^{l-1}_{s}, Q(n)=W^{q}_{\tau(n)}h^{l-1}_{n}
    \end{aligned}
\end{equation}
where $W^{k}_{\tau(s)}\in \mathbb{R}^{d\times d}$ and $W^{q}_{\tau(n)}\in \mathbb{R}^{d\times d}$ are two learnable matrixes.

 



As we use different types of edges for node connections, we follow \newcite{hu2020heterogeneous} to keep a distinct edge-matrix $W_{\phi(e)}\in \mathbb{R}_{d\times d}$ for each edge type between the dot of $Q(n)$ and $K(s)$:
\begin{equation}
    \begin{aligned}
    	&\text{P}(n,s)=K(s)W_{\phi(e)}Q(n)^{\mathsf{T}}\cdot\frac{\mu\langle\tau(s),\phi(e),\tau(n)\rangle}{\sqrt{d}}, \\
      	&\text{A}(s,e,n)=\mathop{\text{softmax}}\limits_{s\in \mathcal{N}(n),e\in \phi(e)}(P(Q(n),K(s))), \\
    \end{aligned}
\end{equation}
where $\mu\in \mathbb{R}^{\vert\mathcal{A}\vert\times\vert\mathcal{R}\vert\times\vert\mathcal{A}\vert}$ is a learnable matrix denoting the contribution of each edge with a different relationship. 
Similar to \newcite{vaswani2017attention}, the final attention is the concatenation of heads which compute the attention weights independently:
\begin{equation}
    \begin{aligned}
    	&\text{MultiHead}(s,e,n)=\text{Concat}(\text{head}_{1}\cdots\text{head}_{g})W^{O},\\
    	&\text{head}_{i}=\text{A}(s,e,n)\cdot\text{M}(s,e,n)
    \end{aligned}
\end{equation}
where $g$ is the number of heads for multi-head attention (8 is used for our model) and $W^{O}$ is a learnable model parameter.

\paragraph{Aggregate} For each edge $(s,e,n)$, we now have the attention weight $\text{A}(s,e,n)$ and the information $\text{M}(s,e,n)$, the next step is to obtain the weighted-sum information from all neighboring nodes:
\begin{equation}
    \begin{aligned}
    	\text{Aggregate}(\cdot)=W^{o}_{\tau(n)}(\mathop{\oplus}\limits_{\forall s\in \mathcal{N}(n)}\text{MultiHead}(s,e,n))
	\end{aligned}
\end{equation}
where $\oplus$ is element-wise addition and $W^{o}_{\tau(n)}\in \mathcal{R}^{d\times d}$ is a learnable model parameter used as an activation function like a linear layer.

The aggregated representation for each input word is used as its final representation, passed to the softmax layer for classification. For all our experiments, the number of heads is 8.

\begin{table}[th!]
    \centering
    \resizebox{.5\textwidth}{!}{
    \begin{tabular}{llll}\toprule
        \multicolumn{4}{c}{{\bf English CoNLL 2003}} \\\midrule
        \textbf{Model} & \textbf{Precision} & \textbf{Recall} & \textbf{F1} \\\midrule
        CVT~ \citep{clark2018semi} & - & - & 92.22 \\
        BERT-MRC~ \citep{li2019unified} & 92.33 & 94.61 & 93.04 \\\midrule
        BERT-Large~ \citep{devlin2018bert} & - & - & \bf{92.8} \\
        BERT-Large+KNN & 92.90 & 92.88 & \bf{92.86 (+0.06)} \\
        BERT-Large+GNN & 92.92 & 93.34 & \bf{93.14 (+0.33)} \\
        BERT-Large+GNN+KNN & 92.95 & 93.37 & \bf{93.16 (+0.35)} \\\midrule
        RoBERTa-Large~ \citep{liu2019roberta} & 92.77 & 92.81 & \bf{92.76} \\
        RoBERTa-Large+KNN & 92.82 & 92.99 & \bf{92.93 (+0.17)} \\
        RoBERTa-Large+GNN & 93.00 & 93.41 & \bf{93.20 (+0.44)} \\
        RoBERTa-Large+GNN+KNN & 93.02 & 93.40 & \bf{93.20 (+0.44)} \\\bottomrule
        \multicolumn{4}{c}{{\bf English OntoNotes 5.0}} \\\midrule
        \textbf{Model} & \textbf{Precision} & \textbf{Recall} & \textbf{F1} \\\midrule
        CVT~ \citep{clark2018semi} & - & - & 88.8 \\
        BERT-MRC~ \citep{li2019unified} & 92.98 & 89.95 & 91.11 \\\midrule
        BERT-Large~ \citep{devlin2018bert} & 90.01 & 88.35 & \bf{89.16} \\
        BERT-Large+KNN & 89.93 & 91.65 & \bf{90.78 (+1.62)} \\
        BERT-Large+GNN & 91.44 & 91.16 & \bf{91.30 (+2.14)} \\
        BERT-Large+GNN+KNN & 91.47 & 91.14 & \bf{91.32 (+2.16)} \\\midrule
        RoBERTa-Large~ \citep{liu2019roberta} & 89.77 & 89.27 & \bf{89.52} \\
        RoBERTa-Large+KNN & 90.00 & 91.26 & \bf{90.63 (+1.11)} \\
        RoBERTa-Large+GNN & 91.38 & 91.17 & \bf{91.30 (+1.78)} \\
        RoBERTa-Large+GNN+KNN & 91.48 & 91.29 & \bf{91.39 (+1.87)} \\\bottomrule
        \multicolumn{4}{c}{{\bf Chinese OntoNotes 4.0}} \\\midrule
        \textbf{Model} & \textbf{Precision} & \textbf{Recall} & \textbf{F1} \\\midrule
        Lattice-LSTM~ \citep{zhang2018chinese} & 76.35 & 71.56 & 73.88 \\
        Glyce-BERT~ \citep{meng2019glyce} & 81.87 & 81.40 & 80.62 \\
        BERT-MRC~ \citep{li2019unified} & 82.98 & 81.25 & 82.11 \\\midrule
        BERT-Large~ \citep{devlin2018bert} & 78.01 & 80.35 & \bf{79.16} \\
        BERT-Large+KNN & 80.23 & 81.60 & \bf{80.91 (+1.75)} \\
        BERT-Large+GNN & 83.06 & 81.60 & \bf{82.33 (+3.17)} \\
        BERT-Large+GNN+KNN & 83.07 & 81.62 & \bf{82.35 (+3.19)} \\\midrule
        ChineseBERT-Large~ \citep{sun2021chinesebert} & 80.77 & 83.65 & \bf{82.18} \\
        ChineseBERT-Large+KNN & 81.68 & 83.46 & \bf{82.56 (+0.38)} \\
        ChineseBERT-Large+GNN & 82.02 & 84.01 & \bf{83.02 (+0.84)} \\
        ChineseBERT-Large+GNN+KNN & 82.21 & 83.98 & \bf{83.10 (+0.92)} \\\bottomrule
        \multicolumn{4}{c}{{\bf Chinese MSRA}} \\\midrule
        \textbf{Model} & \textbf{Precision} & \textbf{Recall} & \textbf{F1} \\\midrule
        Lattice-LSTM~ \citep{zhang2018chinese} & 93.57 & 92.79 & 93.18 \\
        Glyce-BERT~ \citep{meng2019glyce} & 95.57 & 95.51 & 95.54 \\
        BERT-MRC~ \citep{li2019unified} & 96.18 & 95.12 & 95.75 \\\midrule
        BERT-Large~ \citep{devlin2018bert} & 94.97 & 94.62 & \bf{94.80} \\
        BERT-Large+KNN & 95.34 & 94.64 & \bf{94.99 (+0.19)} \\
        BERT-Large+GNN & 96.29 & 95.51 & \bf{95.90 (+1.10)} \\
        BERT-Large+GNN+KNN & 96.31 & 95.54 & \bf{95.93 (+1.13)} \\\midrule
        ChineseBERT-Large~ \citep{sun2021chinesebert} & 95.61 & 95.61 & \bf{95.61} \\
        ChineseBERT-Large+KNN & 95.83 & 95.68 & \bf{95.76 (+0.15)} \\
        ChineseBERT-Large+GNN & 96.28 & 95.73 & \bf{96.01 (+0.40)} \\
        ChineseBERT-Large+GNN+KNN & 96.29 & 95.75 & \bf{96.03 (+0.42)} \\\bottomrule
    \end{tabular}
    }
     \caption{NER results for two English datasets: CoNLL 2003 and OntoNotes 5.0, and two Chinese datasets: MSRA and OntoNotes 4.0.}
    \label{tab:ner_result}
\end{table}


\begin{table}[th!]
    \centering
    \resizebox{.5\textwidth}{!}{
    \begin{tabular}{llll}\toprule
        \multicolumn{4}{c}{{\bf PKU}} \\\midrule
        \textbf{Model} & \textbf{Precision} & \textbf{Recall} & \textbf{F1} \\\midrule
        Multitask pretrain~ \citep{yang2017neural} & - & - & 96.3 \\
        CRF-LSTM~ \citep{huang2019toward} & - & - & 96.6 \\
        Glyce-BERT~ \citep{meng2019glyce} & 97.1 & 96.4 & 96.7 \\\midrule
        BERT-Large~ \citep{devlin2018bert} & 96.8 & 96.3 & \bf{96.5} \\
        BERT-Large+KNN & 97.2 & 96.1 & \bf{96.6 (+0.1)} \\
        BERT-Large+GNN & 96.9 & 96.6 & \bf{96.8 (+0.3)} \\
        BERT-Large+GNN+KNN & 96.9 & 96.7 & \bf{96.8 (+0.3)} \\\midrule
        ChineseBERT-Large~ \citep{sun2021chinesebert} & 97.3 & 96.0 & \bf{96.7} \\
        ChineseBERT-Large+KNN & 97.3 & 96.1 & \bf{96.7 (+0.0)} \\
        ChineseBERT-Large+GNN & 97.6 & 96.2 & \bf{96.9 (+0.2)} \\
        ChineseBERT-Large+GNN+KNN & 97.7 & 96.2 & \bf{96.9 (+0.2)} \\\bottomrule
        \multicolumn{4}{c}{{\bf CITYU}} \\\midrule
        \textbf{Model} & \textbf{Precision} & \textbf{Recall} & \textbf{F1} \\\midrule
        Multitask pretrain~ \citep{yang2017neural} & - & - & 96.9 \\
        CRF-LSTM~ \citep{huang2019toward} & - & - & 97.6 \\
        Glyce-BERT~ \citep{meng2019glyce} & 97.9 & 98.0 & 97.9 \\\midrule
        BERT-Large~ \citep{devlin2018bert} & 97.5 & 97.7 & \bf{97.6} \\
        BERT-Large+KNN & 97.8 & 97.8 & \bf{97.8 (+0.2)} \\
        BERT-Large+GNN & 98.0 & 98.1 & \bf{98.0 (+0.4)} \\
        BERT-Large+GNN+KNN & 98.0 & 98.1 & \bf{98.0 (+0.4)} \\\midrule
        ChineseBERT-Large~ \citep{sun2021chinesebert} & 97.8 & 98.2 & \bf{98.0} \\
        ChineseBERT-Large+KNN & 98.1 & 98.0 & \bf{98.1 (+0.1)} \\
        ChineseBERT-Large+GNN & 98.2 & 98.4 & \bf{98.3 (+0.3)} \\
        ChineseBERT-Large+GNN+KNN & 98.3 & 98.4 & \bf{98.3 (+0.3)} \\\bottomrule
        \multicolumn{4}{c}{{\bf MSR}} \\\midrule
        \textbf{Model} & \textbf{Precision} & \textbf{Recall} & \textbf{F1} \\\midrule
        Multitask pretrain~ \citep{yang2017neural} & - & - & 97.5 \\
        CRF-LSTM~ \citep{huang2019toward} & - & - & 97.9 \\
        Glyce-BERT~ \citep{meng2019glyce} & 98.2 & 98.3 & 98.3 \\\midrule
        BERT-Large~ \citep{devlin2018bert} & 98.1 & 98.2 & \bf{98.1} \\
        BERT-Large+KNN & 98.3 & 98.4 & \bf{98.3 (+0.2)} \\
        BERT-Large+GNN & 98.4 & 98.3 & \bf{98.4 (+0.3)} \\
        BERT-Large+GNN+KNN & 98.4 & 98.3 & \bf{98.4 (+0.3)} \\\midrule
        ChineseBERT-Large~ \citep{sun2021chinesebert} & 98.5 & 98.0 & \bf{98.3} \\
        ChineseBERT-Large+KNN & 98.5 & 98.1 & \bf{98.3 (+0.0)} \\
        ChineseBERT-Large+GNN & 98.9 & 97.9 & \bf{98.5 (+0.2)} \\
        ChineseBERT-Large+GNN+KNN & 98.9 & 98.0 & \bf{98.5 (+0.2)} \\\bottomrule
        \multicolumn{4}{c}{{\bf AS}} \\\midrule
        \textbf{Model} & \textbf{Precision} & \textbf{Recall} & \textbf{F1} \\\midrule
        Multitask pretrain~ \citep{yang2017neural} & - & - & 95.7 \\
        CRF-LSTM~ \citep{huang2019toward} & - & - & 96.6 \\
        Glyce-BERT~ \citep{meng2019glyce} & 96.6 & 96.8 & 96.7 \\\midrule
        BERT-Large~ \citep{devlin2018bert} & 96.7 & 96.4 & \bf{96.5} \\
        BERT-Large+KNN & 96.2 & 96.9 & \bf{96.6 (+0.1)} \\
        BERT-Large+GNN & 96.6 & 97.0 & \bf{96.8 (+0.3)} \\
        BERT-Large+GNN+KNN & 96.6 & 97.0 & \bf{96.8 (+0.3)} \\\midrule
        ChineseBERT-Large~ \citep{sun2021chinesebert} & 96.3 & 97.2 & \bf{96.7} \\
        ChineseBERT-Large+KNN & 96.3 & 97.2 & \bf{96.7 (+0.0)} \\
        ChineseBERT-Large+GNN & 96.1 & 97.7 & \bf{96.9 (+0.2)} \\
        ChineseBERT-Large+GNN+KNN & 96.2 & 97.7 & \bf{96.9 (+0.2)} \\\bottomrule
    \end{tabular}
    }
     \caption{CWS results for four datasets: PKU, CITYU, MSR, and AS.}
    \label{tab:cws_result_on}
\end{table}

\begin{table*}[th!]
    \centering
    \resizebox{\textwidth}{!}{
    \begin{tabular}{llllllllll}\toprule
        & \multicolumn{3}{c}{{\bf Chinese CTB5}} & \multicolumn{3}{c}{{\bf Chinese CTB6}} & \multicolumn{3}{c}{{\bf Chinese UD1.4}} \\\midrule
        \textbf{Model} & \textbf{Precision} & \textbf{Recall} & \textbf{F1} & \textbf{Precision} & \textbf{Recall} & \textbf{F1} & \textbf{Precision} & \textbf{Recall} & \textbf{F1} \\\midrule
        Joint-POS(Sig)~ \citep{shao2017character} & 93.68 & 94.47 & 94.07 & - & - & 90.81 & 89.28 & 89.54 & 89.41 \\
        Joint-POS(Ens)~ \citep{shao2017character} & 93.95 & 94.81 & 94.38 & - & - & - & 89.67 & 89.86 & 89.75 \\
       	Lattice-LSTM~ \citep{zhang2018chinese} & 94.77 & 95.51 & 95.14 & 92.00 & 90.86 & 91.43 & 90.47 & 89.70 & 90.09 \\
        Glyce-BERT~ \citep{meng2019glyce} & 96.50 & 96.74 & 96.61 & 95.56 & 95.26 & 95.41 & 96.19 & 96.10 & 96.14 \\\midrule
        BERT-Large~ \citep{devlin2018bert} & 95.86 & 96.26 & \bf{96.06} & 94.91 & 94.63 & \bf{94.77} & 95.42 & 94.17 & \bf{94.79} \\
        BERT-Large+KNN & 96.36 & 96.60 & \bf{96.48 (+0.42)} & 95.14 & 94.77 & \bf{94.95 (+0.18)} & 95.85 & 95.67 & \bf{95.76 (+0.97)} \\
        BERT-Large+GNN & 96.86 & 96.66 & \bf{96.76 (+0.70)} & 96.46 & 94.82 & \bf{95.64 (+0.85)} & 96.14 & 96.46 & \bf{96.30 (+1.51)} \\
        BERT-Large+GNN+KNN & 96.77 & 96.69 & \bf{96.79 (+0.73)} & 96.44 & 94.84 & \bf{95.64 (+0.85)} & 96.20 & 96.47 & \bf{96.34 (+1.55)} \\\midrule
        ChineseBERT~ \citep{sun2021chinesebert} & 96.35 & 96.54 & \bf{96.44} & 95.47 & 95.00 & \bf{95.23} & 96.02 & 95.92 & \bf{95.97} \\
        ChineseBERT+KNN & 96.41 & 96.15 & \bf{96.52 (+0.08)} & 95.48 & 95.09 & \bf{95.29 (+0.06)} & 96.11 & 96.11 & \bf{96.11 (+0.14)} \\
        ChineseBERT+GNN & 96.46 & 97.41 & \bf{96.94 (+0.50)} & 96.08 & 95.50 & \bf{95.79 (+0.77)} & 96.18 & 96.54 & \bf{96.36 (+0.39)} \\
        ChineseBERT+GNN+KNN & 96.46 & 97.44 & \bf{96.96 (+0.52)} & 96.13 & 95.58 & \bf{95.85 (+0.82)} & 96.25 & 96.57 & \bf{96.40 (+0.43)} \\\bottomrule
    \end{tabular}
    }
     \caption{POS results for three Chinese datasets: CTB5, CTB6 and UD1.4.}
    \label{tab:pos_result_on_chinese}
\end{table*}

\begin{table}[th!]
    \centering
    \resizebox{.5\textwidth}{!}{
    \begin{tabular}{llll}\toprule
        \multicolumn{4}{c}{{\bf English WSJ}} \\\midrule
        \textbf{Model} & \textbf{Precision} & \textbf{Recall} & \textbf{F1} \\\midrule
        Meta BiLSTM~ \citep{bohnet2018morphosyntactic} & - & - & 98.23 \\\midrule
        BERT-Large~ \citep{devlin2018bert} & 99.21 & 98.36 & \bf{98.86} \\
        BERT-Large+KNN & 98.98 & 98.85 & \bf{98.92 (+0.06)} \\
        BERT-Large+GNN & 98.84 & 98.98 & \bf{98.94 (+0.08)} \\
        BERT-Large+GNN+KNN & 98.88 & 98.99 & \bf{98.96 (+0.10)} \\\midrule
        RoBERTa-Large~ \citep{liu2019roberta} & 99.22 & 98.44 & \bf{98.90} \\
        RoBERTa-Large+KNN & 99.21 & 98.52 & \bf{98.94 (+0.04)} \\
        RoBERTa-Large+GNN & 98.90 & 99.06 & \bf{99.00 (+0.10)} \\
        RoBERTa-Large+GNN+KNN & 98.90 & 99.06 & \bf{99.00 (+0.10)} \\\bottomrule
        \multicolumn{4}{c}{{\bf English Tweets}} \\\midrule
        \textbf{Model} & \textbf{Precision} & \textbf{Recall} & \textbf{F1} \\\midrule
        FastText+CNN+CRF & - & - & 91.78 \\\midrule
        BERT-Large~ \citep{devlin2018bert} & 92.33 & 91.98 & \bf{92.34} \\
        BERT-Large+KNN & 92.77 & 92.02 & \bf{92.39 (+0.05)} \\
        BERT-Large+GNN & 92.38 & 92.52 & \bf{92.45 (+0.11)} \\
        BERT-Large+GNN+KNN & 92.42 & 92.53 & \bf{92.48 (+0.14)} \\\midrule
        RoBERTa-Large~ \citep{liu2019roberta} & 92.40 & 91.99 & \bf{92.38} \\
        RoBERTa-Large+KNN & 92.44 & 92.11 & \bf{92.46 (+0.08)} \\
        RoBERTa-Large+GNN & 92.49 & 92.53 & \bf{92.51 (+0.13)} \\
        RoBERTa-Large+GNN+KNN & 92.49 & 92.54 & \bf{92.52 (+0.14)} \\\bottomrule
    \end{tabular}
    }
     \caption{POS results for two English datasets: WSJ and Tweets.}
    \label{tab:pos_result_on_english}
\end{table}

\section{Experiments}
We conduct experiments on three widely-used sub-tasks of sequence labeling: named entity recognition (NER), part of speech tagging (POS), and Chinese Word Segmentation (CWS). 

\subsection{Trainng Details}
\paragraph{The Vanilla SL Model}
\label{vanilla_model_detail}
As described in Section \ref{proposed_method}, we need the pre-trained vanilla SL model to extract features to initial the nodes of the constructed graph. For all our experiments, we choose the standard BERT-large \cite{devlin2018bert} and RoBERTa-large \cite{liu2019roberta} for English tasks, as well as the standard BERT-large and ChineseBERT-large \cite{sun2021chinesebert} for Chinese tasks.

\paragraph{$k$NN Retrieval}
In the process of $k$NN retrieval, 
the number of nearest neighbors $k$ is set to 32, and the size of the nearest context window is set to 7 (setting both the left and right side of the window to 3). The two numbers are chosen according to the evaluation in Section \ref{ablation_study} and perform best in our experiments. For the $k$ nearest search, we use the last layer output of the pre-trained vanilla SL model as the representation and the $L^{2}$ distance as the metric of similarity comparison.

\subsection{Control Experiments}
\label{control_experiments}
To better show the effectiveness of the proposed model, we compare the performance of the following setups:
(1) {\bf vanilla SL models}:  vanilla models naturally constitute a  baseline for comparison, where the final layer representation is fed to a softmax function
to obtain $p_{\text{vanilla}}$
 for label prediction; 
(2) {\bf vanilla +  $k$NN}:  the $k$NN probability $p_{\text{kNN}}$ is interpolated with  $p_{\text{vanilla}}$  to obtain final predictions;
(3) {\bf vanilla +  GNN}: the representation generated from the final layer of GNN is passed to the softmax layer to obtain the label probability $p_{\text{GNN}}$;
(4) {\bf vanilla +  GNN +  $k$NN}: the $k$NN probability $p_{\text{kNN}}$ is interpolated with the GNN probability $p_{\text{GNN}}$ to obtain final predictions, rather than the probability from the vanilla model, as in vanilla +  $k$NN.

\begin{table*}[th!]
    \centering
    \resizebox{\textwidth}{!}{
    \begin{tabular}{lc}\toprule
    	\textbf{Input Sentence \#1} & \\\midrule
    	\text{Hornak moved on from Tigers to \underline{\it Phoenix} for studies and work.} & \\
    	\text{{\bf Ground Truth:} ORGANIZATION, {\bf Vanilla SL Output:} LOCATION, {\bf GNN-SL Output:} ORGANIZATION} & \\\midrule
        \textbf{Retrieved Nearest Neighbors} & \textbf{Retrieved Label} \\\midrule
        \text{{\bf 1-th}: Hornak signed accomplished performance in a Tigers display against \underline{\it Phoenix}.} & \text{ORGANIZATION} \\
        \text{{\bf 8-th}: Blinker was fined 75,000 Swiss francs (\$57,600) for failing to inform the English club of his previous commitment to \underline{\it Udinese}.} & \text{ORGANIZATION} \\
        \text{{\bf 16-th}: Since we have friends in \underline{\it Phoenix}, we pop in there for a brief visit.} & \text{LOCATION} \\\bottomrule
    	\textbf{Input Sentence \#2} & \\\midrule
        \text{Australian \underline{\it Tom Moody} took six for 82 but Tim O'Gorman, 109, took Derbyshire to 471.} & \\
        \text{{\bf Ground Truth:} PERSON, {\bf Vanilla SL Output:} - (Not an Entity), {\bf GNN-SL Output:} PERSON} & \\\midrule
        \textbf{Retrieved Nearest Neighbors} & \textbf{Retrieved Label} \\\midrule
        \text{{\bf 1-th}: At California, \underline{\it Troy O'Leary} hit solo home runs in the second inning as the surging Boston Red Sox.} & \text{PERSON} \\
        \text{{\bf 8-th}: Britain's \underline{\it Chris Boardman} broke the world 4,000 meters cycling record by more than six seconds.} & \text{PERSON} \\
        \text{{\bf 16-th}: Japan coach \underline{\it Shu Kamo} said: The Syrian own goal proved lucky for us.} & \text{PERSON} \\\bottomrule
    \end{tabular}
    }
     \caption{Retrieved nearest examples from English OntoNotes 5.0 dataset, where the labeled words are underlined.}
    \label{tab:examples}
\end{table*}


\subsection{Named Entity Recognition}
\paragraph{Details}
The task of NER is normally treated as a char-level tagging task: outputting a NER tag for each character.
We conduct experiments on CoNLL2003 \cite{sang2003introduction} and OntoNotes5.0 \cite{pradhan2013towards} for English, and MSRA \cite{levow-2006-third}, OntoNotes4.0 \cite{pradhan2011proceedings} for Chinese.

For the chosen baselines, we make a comparison with Cross-View Training CVT \cite{clark2018semi}, BERT-MRC \cite{li2019unified} for English datasets and Lattice-LSTM \cite{zhang2018chinese}, Glyce-BERT \cite{meng2019glyce}, BERT-MRC \cite{li2019unified} for Chinese datasets.
\paragraph{Results}
\label{ner_results}
Results for the NER task are shown in Table \ref{tab:ner_result}, and from the results: 

(1) 
We observe a significant performance boost brought by $k$NN, respectively \text{+0.06}, \text{+1.62}, \text{+1.75} and \text{+0.19} for English CoNLL 2003, English OntoNotes 5.0, Chinese OntoNotes 4.0 and Chinese MSRA. 
which proves the importance of incorporating the evidence of retrieved neighbors. 

(2) We observe a significant performance boost for vanilla+GNN over both the vanilla model:
respectively \text{+2.14} on the BERT-Large and \text{+1.78} on the RoBERTa-Large for English OntoNotes 5.0, and \text{+1.10} on the BERT-Large and \text{+0.40} on the ChineseBERT-Large for Chinese MSRA.
Due to the fact that both vanilla+GNN and the vanilla model output the final layer representation to the softmax function to obtain final probability and that $p_{\text{kNN}}$ do not participate in the final probability interpolation for both,
the performance boost over vanilla SL demonstrates that we are able to obtain better token-level representations using GNNs. 

(3)
When comparing with  vanilla+$k$NN, 
we observe further improvements of \text{+0.33}, \text{+2.14}, \text{+3.17} and \text{+1.10} for English CoNLL 2003, English OntoNotes 5.0, Chinese OntoNotes 4.0 and Chinese MSRA dataset, 
respectively, 
showing that the proposed GNN model has the ability to filtrate the useful neighbors to augment the vanilla SL model;

(4) There is an inconspicuous improvement brought by interpolating both the $k$NN probability and GNN probability (vanilla + $k$NN + GNN) over the proposed GNN-SL (vanilla + GNN), e.g., \text{+2.16} v.s. \text{+2.14} on the BERT-Large for English OntoNotes 5.0, and \text{+1.13} v.s. \text{+1.10} on the BERT-Large for Chinese MSRA.
This 
demonstrates that as the evidence of retrieved nearest neighbors (and their labels) has been assimilated through GNNs in the representation learning stage,
the extra benefits brought by interpolating $p_{\text{knn}}$ in the final prediction stage is significantly narrowed. 

\subsection{Chinese Word Segmentation}
\paragraph{Details}
The task of CWS is normally treated as a char-level tagging problem: assigning \textit{seg} or \textit{not seg} for each input word.

For evaluation, four Chinese datasets retrieved from SIGHAN 2005\footnote{The website of the 4-th Second International Chinese Word Segmentation Bakeoff (SIGHAN 2005) is: http://sighan.cs.uchicago.edu/bakeoff2005/} are used: PKU, MSR, CITYU, and AS, and benchmarks Multitask pretrain \cite{yang2017neural}, CRF-LSTM \cite{huang2019toward} and Glyce+BERT \cite{meng2019glyce} are chosen for the comparison.

\paragraph{Results}
Results for the CWS task are shown in Table \ref{tab:cws_result_on}. From the results, same as the former Section \ref{ner_results}, with different vanilla models as the backbone, we can observe obvious improvements by applying the $k$NN probability (vanilla + $k$NN) or the GNN model (vanilla + GNN), while keeping the same results between vanilla + GNN and vanilla + GNN + $k$NN, e.g., for PKU dataset \text{+0.1} on BERT + $k$NN, \text{+0.3} both on BERT + GNN and BERT + GNN + $k$NN. Notablely we achieve SOTA for all four datasets with the ChineseBERT: \text{96.9 (+0.2)} on PKU, \text{98.3 (+0.3)} on CITYU, \text{98.5 (+0.2)} on MSR and \text{96.9 (+0.2)} on AS.

\subsection{Part of Speech Tagging}
\paragraph{Details}
The task of POS is normally formalized as a character-level sequence labeling task, assigning labels to each of the input word.

We use Wall Street Journal (WSJ) and Tweets \cite{ritter-etal-2011-named} for English datasets, Chinese Treebank 5.0, Chinese Treebank 6.0, and UD1.4 \cite{xue2005penn} for Chinese datasets.

For the comparisons, we choose Meta-BiLSTM \cite{bohnet2018morphosyntactic} for English datasets and Joint-POS \cite{shao2017character}, Lattice-LSTM \cite{zhang2018chinese} and Glyce-BERT \cite{meng2019glyce} for Chinese datasets.

\paragraph{Results} Results for the POS task are shown in Table \ref{tab:pos_result_on_chinese} for Chinese datasets and Table \ref{tab:pos_result_on_english} for English datasets. As shown, with different vanilla models and datasets, the phenomenons are the same as the former Section \ref{ner_results} that improves largely based on the $k$NN probability and further on the GNN model. For the results of Chinese CTB5 as the example, \text{+0.42} on BERT + $k$NN, further \text{+0.70} on BERT + GNN, and \text{+0.73} on BERT + GNN + $k$NN. 

\paragraph{Examples} To illustrate the augment of our proposed GNN-SL, we visualize the retrieved $k$NN examples as well as the input sentence in Table \ref{tab:examples}. For the first example the long-tail case “\text{\it Phoenix}”, which is assigned with “\text{\it LOCATION}” by the vanilla SL model, is amended to “\text{\it ORGANIZATION}” by the nearest neighbors. Especially, we can observe that both 1-th and 8-th retrieved labels are “\text{\it ORGANIZATION}” while the 16-th retrieved label is “\text{\it LOCATION}” which is against the ground truth. That phenomenon proves that retrieved neighbors do relate to the input sentence in different ways: some are close to the original input sentence while others are just noise, and our proposed GNN-SL has the ability to better model the relationships between the retrieved nearest examples and the input word. For the second example, with the augment of the retrieved nearest neighbors, our proposed GNN-SL outputs the correct label “\text{\it PERSON}” for the word “\text{\it Tom Moody}”.

\begin{table*}[th!]
    \centering
    \resizebox{\textwidth}{!}{
    \begin{tabular}{lllllllllllll}\toprule
    	\multicolumn{13}{c}{{\it Task NER}} \\
        & \multicolumn{3}{c}{{\bf English CoNLL 2003}} & \multicolumn{3}{c}{{\bf English OntoNotes 5.0}} & \multicolumn{3}{c}{{\bf Chinese OntoNotes 4.0}} & \multicolumn{3}{c}{{\bf Chinese MSRA}} \\\midrule
        \textbf{Model} & \textbf{Precision} & \textbf{Recall} & \textbf{F1} & \textbf{Precision} & \textbf{Recall} & \textbf{F1} & \textbf{Precision} & \textbf{Recall} & \textbf{F1} & \textbf{Precision} & \textbf{Recall} & \textbf{F1} \\\midrule
        BERT+GNN & 92.97 & 93.37 & \bf{93.17} & 91.40 & 91.15 & \bf{91.27} & 83.04 & 81.56 & \bf{82.30} & 96.69 & 95.51 & \bf{95.90} \\
        BERT+GNN-\textit{without Label nodes} & 92.99 & 93.29 & \bf{93.14 (-0.03)} & 91.66 & 90.70 & \bf{91.18 (-0.09)} & 83.23 & 81.06 & \bf{82.13 (-0.17)} & 95.86 & 95.31 & \bf{95.58 (-0.32)} \\\bottomrule
        \multicolumn{13}{c}{{\it Task CWS}} \\
        & \multicolumn{3}{c}{{\bf PKU}} & \multicolumn{3}{c}{{\bf CITYU}} & \multicolumn{3}{c}{{\bf MSR}} & \multicolumn{3}{c}{{\bf AS}} \\\midrule
        \textbf{Model} & \textbf{Precision} & \textbf{Recall} & \textbf{F1} & \textbf{Precision} & \textbf{Recall} & \textbf{F1} & \textbf{Precision} & \textbf{Recall} & \textbf{F1} & \textbf{Precision} & \textbf{Recall} & \textbf{F1} \\\midrule
        BERT+GNN & 96.9 & 96.6 & \bf{96.8} & 98.0 & 98.1 & \bf{98.0} & 98.4 & 98.3 & \bf{98.4} & 96.6 & 97.0 & \bf{96.8} \\
        BERT+GNN-\textit{without Label nodes} & 97.1 & 96.2 & \bf{96.7 (-0.1)} & 97.9 & 97.8 & \bf{97.9 (-0.1)} & 98.4 & 98.2 & \bf{98.3 (-0.1)} & 96.3 & 97.1 & \bf{96.7 (-0.1)} \\\bottomrule
        \multicolumn{13}{c}{{\it Task POS}} \\
        & \multicolumn{3}{c}{{\bf Chinese CTB5}} & \multicolumn{3}{c}{{\bf Chinese CTB6}} & \multicolumn{3}{c}{{\bf Chinese UD1.4}} & \multicolumn{3}{c}{{\bf English WSJ}} \\\midrule
        \textbf{Model} & \textbf{Precision} & \textbf{Recall} & \textbf{F1} & \textbf{Precision} & \textbf{Recall} & \textbf{F1} & \textbf{Precision} & \textbf{Recall} & \textbf{F1} & \textbf{Precision} & \textbf{Recall} & \textbf{F1} \\\midrule
        BERT+GNN & 96.87 & 96.69 & \bf{96.78} & 96.44 & 94.82 & \bf{95.63} & 96.15 & 96.47 & \bf{96.31} & 98.85 & 98.99 & \bf{98.95} \\
        BERT+GNN-\textit{without Label nodes} & 96.36 & 96.58 & \bf{96.46 (-0.32)} & 95.69 & 94.69 & \bf{95.19 (-0.42)} & 95.99 & 96.05 & \bf{96.02 (-0.29)} & 98.97 & 98.87 & \bf{98.92 (-0.003)} \\\bottomrule
    \end{tabular}
    }
     \caption{Experiments without label nodes on three tasks: NER, CWS, and POS.}
    \label{tab:ablation_label_node}
\end{table*}

\begin{figure}[htb]
    \includegraphics[scale=0.37]{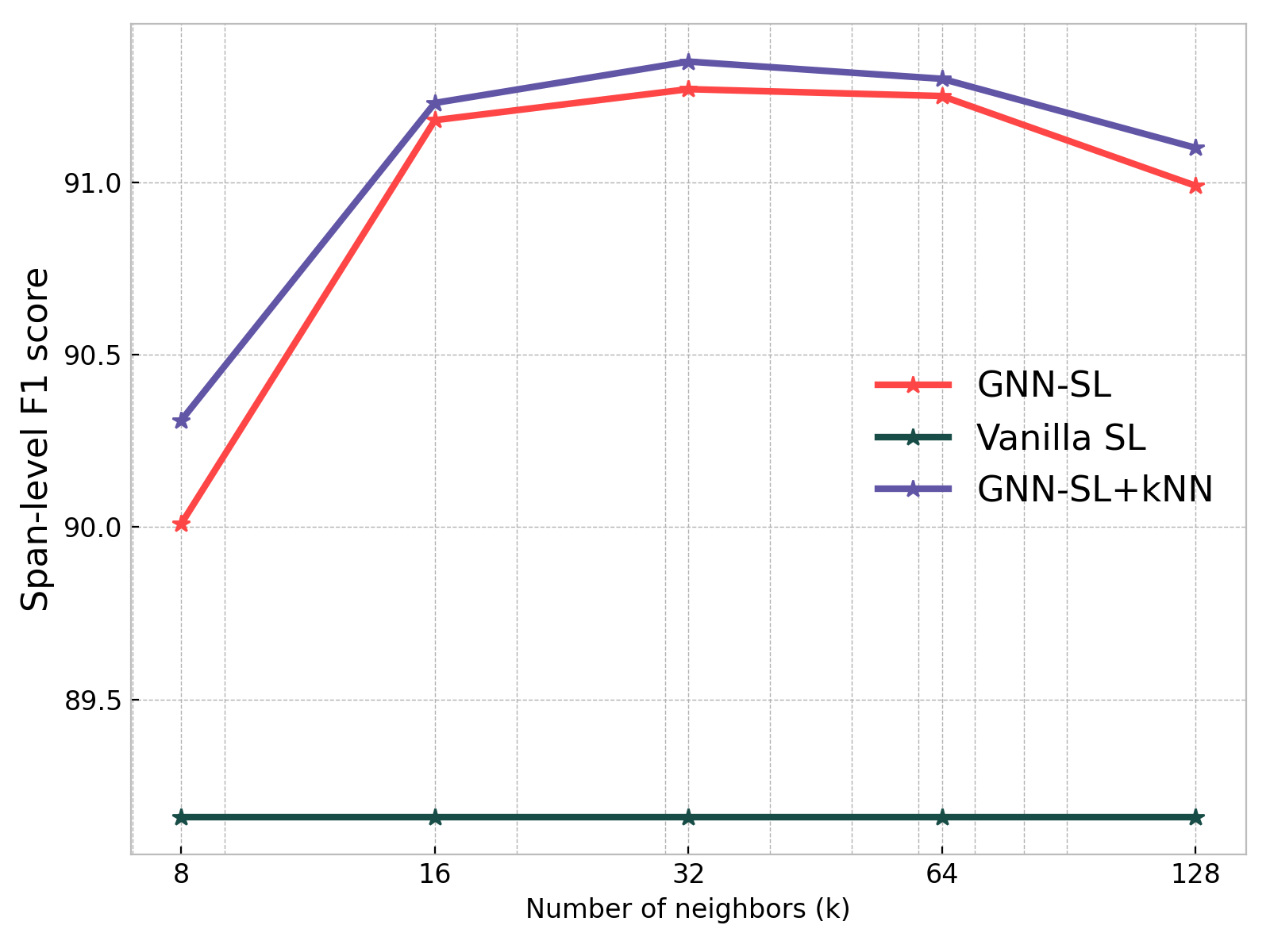}
    \caption{Experiments on English OntoNotes 5.0 datasets by varying the number of neighbors $k$.}
    \label{fig:ablation_gnn_knn}
\end{figure}

\begin{table}[th!]
    \tiny
    \centering
    \resizebox{.405\textwidth}{!}{
    \begin{tabular}{ll}\toprule
        \multicolumn{2}{c}{{\bf F1-score on English OntoNotes 5.0}} \\\midrule
        \textbf{Context (l+r+1) / Model} & \textbf{F1-score} \\\midrule
        \text{The Vanilla SL Model} & 89.16 \\\midrule
       	\text{GNN-SL} &  \\
         \text{+ by setting context=3} & 91.16 (+2.00) \\
         \text{+ by setting context=5} & 91.24 (+2.08) \\
         \text{+ by setting context=7} & 91.27 (+2.11) \\
         \text{+ by setting context=9} & 91.27 (+2.11) \\
         \text{+ by setting context=11} & 91.27 (+2.11) \\\bottomrule
    \end{tabular}
    }
     \caption{F1-score on English OntoNotes 5.0 by varying the context size of the retrieved neighbors}
    \label{tab:context_l_r}
\end{table}

\section{Ablation Study}
\label{ablation_study}

\subsection{The Number of Retrieved Neighbors} 
To evaluate the 
 influence of the amount of the retrieved information from the training set, we conduct experiments 
 on English OntoNotes 5.0 for the NER task
  by varying the number of neighbors $k$. The results are shown in Figure \ref{fig:ablation_gnn_knn}.
 As can be seen, as $k$ increases, 
  the F1 score of GNN-SL first increases and then decreases. The explanation is as follows, as more examples are, 
  more noise is introduced and
  relevance to the query decreases, which makes performance worse.

\subsection{Effectiveness of Label Nodes} 
In Section \ref{building_graph}, labels are used as nodes in the graph construction process, where their influence can be propagated to the input nodes through 
$r_{\text{label-neighbor}}$ and $r_{\text{neighbor-input}}$. 
To evaluate the effectiveness of that strategy, we conducted contrast experiments by removing the label nodes. The experiments are based on the BERT-Large model and adjusted to the best parameters, and the results are shown in Table \ref{tab:ablation_label_node} for all three tasks. All the results show a decrease after removing the label nodes, especially \text{-0.42} for the POS Chinese CTB6 dataset and \text{-0.32} for the NER Chinese MSRA dataset, which proves the necessity of 
directly incorporating label information of neighbors in passing the message. 

\subsection{The Size of the Context Window}
In Section \ref{building_graph}, to acquire the context information of each retrieved nearest word we expand the retrieved nearest word to the nearest context.  We experiment with 
varying
 context sizes to show the influence. Results 
 on English OntoNotes 5.0 for the NER task 
 are shown in Table \ref{tab:context_l_r}.
 We can observe that, as the context size increases, performance first 
goes up and then plateaus. 
The explanation is as follows: a decent size of context is sufficient to provide enough information for predictions. 


\subsection{$k$NN search without Fine-tuning} 
In section \ref{proposed_method}, we use 
the representations obtained by
the fine-tuned pre-trained model on the labeled training set to perform $k$NN search. 
To validate its necessity, we also conduct an experiment that 
 directly uses the BERT model without fine-tuning to extract representation.
 We evaluate its influence on CoNLL 2003 for the NER task 
 and observe a sharp decrease when switching the fine-tuned SL model to a non fine-tuned BERT model, i.e., 93.17 v.s. 92.83. 
 The explanation is due to the gap between the language model task and the sequence labeling task, and that 
 nearest neighbors retrieved by a vanilla pre-trained language model might not be the \textsc{nearest} neighbor for the SL task.

\section{Conclusion} In this work, we propose graph neural networks sequence labeling (GNN-SL), which augments the vanilla SL model output with similar tagging examples retrieved from the whole training set. Since not all the retrieved tagging examples benefit the model prediction, we construct a heterogeneous graph, in which nodes represent the input words associating its nearest examples and edges represent the relationship between each node, and leverage graph neural networks (GNNs) to transfer information from the retrieved nearest tagging examples to the input word. This strategy enables the model to directly acquire similar tagging examples and improves the effectiveness in handling long-tail cases. We conduct multi experiments and analyses on three sequence labeling tasks: NER, POS, and CWS. Notably, GNN-SL achieves SOTA 96.9 (+0.2) on PKU, 98.3 (+0.4) on CITYU, 98.5 (+0.2) on MSR, and 96.9 (+0.2) on AS for the CWS task.

\bibliography{anthology,custom}
\bibliographystyle{acl_natbib}

\newpage
\appendix


\end{document}